\documentclass[letterpaper, 10 pt, conference]{ieeeconf}
\IEEEoverridecommandlockouts

\usepackage[colorlinks]{hyperref}
\hypersetup{
    colorlinks,
    linkcolor=black,
    citecolor=black,
    filecolor=magenta,
    urlcolor=cyan,
}

\usepackage{resizegather}

\usepackage{cite}
\usepackage{amsmath,amssymb,amsfonts,mathrsfs}
\usepackage{algorithmic}
\usepackage{graphicx}
\usepackage{textcomp}
\usepackage{xcolor}
\usepackage{tcolorbox}

\usepackage{tablefootnote}
\usepackage{threeparttable}
\usepackage{caption, subcaption}

\usepackage{tikz}
\usepackage{graphicx}
\usepackage{tkz-euclide}

\usetikzlibrary{positioning}
\usetikzlibrary{shapes}
\usetikzlibrary{shapes.misc}
\usetikzlibrary{shapes.geometric}
\usetikzlibrary{plotmarks}
\usetikzlibrary{intersections}
\usetikzlibrary{calc}
\usetikzlibrary{fit}
\usetikzlibrary{patterns,tikzmark}
\usetikzlibrary{matrix,decorations.pathreplacing,calc}

\tikzset{cross/.style={cross out, draw, 
         minimum size=2*(#1-\pgflinewidth), 
         inner sep=0pt, outer sep=0pt}}

\newcommand{\state}[0]{x}
\newcommand{\prel}[0]{p_{\rm{rel}}}
\newcommand{\vrel}[0]{v_{\rm{rel}}}
\newcommand{\preldot}[0]{\dot{p}_{\rm{rel}}}
\newcommand{\vreldot}[0]{\dot{v}_{\rm{rel}}}

\newcommand{\norm}[1]{\left\lVert#1\right\rVert}

\newcommand{\vertiii}[1]{{\left\vert\kern-0.25ex\left\vert\kern-0.25ex\left\vert #1 
    \right\vert\kern-0.25ex\right\vert\kern-0.25ex\right\vert}}

\newtheorem{theorem}{Theorem}

\newtheorem{remark}{Remark}

\newtheorem{definition}{Definition}

\makeatletter
\renewcommand{\fps@figure}{htp}
\renewcommand{\fps@table}{htp}
\makeatother

\def\BibTeX{{\rm B\kern-.05em{\sc i\kern-.025em b}\kern-.08em
    T\kern-.1667em\lower.7ex\hbox{E}\kern-.125emX}}
    
\begin{document}

\title{Collision Cone Control Barrier Functions for Kinematic \\ Obstacle Avoidance in UGVs}

\author{Phani Thontepu$^{*1}$, Bhavya Giri Goswami$^{*1}$, Manan Tayal$^{1}$, Neelaksh Singh$^{1}$, Shyam Sundar P I$^{1}$, \\Shyam Sundar M G$^{1}$, Suresh Sundaram$^{1}$, Vaibhav Katewa$^{1}$, Shishir Kolathaya$^{1}$
\thanks{This research was supported by the Wipro IISc Research Innovation Network (WIRIN).
}
\thanks{$^{1}$Robert Bosch Center for Cyber Physical Systems (RBCCPS), Indian Institute of Science (IISc), Bengaluru.
{\tt\scriptsize \{vkatewa, shishirk\}@iisc.ac.in}
$^*$These authors have contributed equally.
}%
}

\maketitle
\begin{abstract}

In this paper, we propose a new class of Control Barrier Functions (CBFs) for Unmanned Ground Vehicles (UGVs) that help avoid collisions with kinematic (non-zero velocity) obstacles. While the current forms of CBFs have been successful in guaranteeing safety/collision avoidance with static obstacles, extensions for the dynamic case have seen limited success. Moreover, with the UGV models like the unicycle or the bicycle, applications of existing CBFs have been conservative in terms of control, i.e., steering/thrust control has not been possible under certain scenarios. Drawing inspiration from the classical use of collision cones for obstacle avoidance in trajectory planning, we introduce its novel CBF formulation with theoretical guarantees on safety for both the unicycle and bicycle models. The main idea is to ensure that the velocity of the obstacle w.r.t. the vehicle is always pointing away from the vehicle. Accordingly, we construct a constraint that ensures that the velocity vector always avoids a cone of vectors pointing at the vehicle. The efficacy of this new control methodology is later verified by Pybullet simulations on TurtleBot3 and F1Tenth.
\end{abstract}


\section{Introduction}
\label{section: Introduction}


\par Advances in autonomy have enabled robot application in all kinds of environments and in close interactions with humans, including autonomous navigation. In dynamic environments, i.e., containing moving obstacles, the collision avoidance system must accommodate an unpredictable information picture providing only a limited time to react to a collision. As a result, the effectiveness of planning-based algorithms is reduced significantly, highlighting the need for developing real-time reactive methods that ensure safety. Thus, designing controllers with formal real-time safety guarantees has become an essential aspect of such safety-critical applications and an active research area in recent years.  Researchers have developed many tools to handle this problem, such as reachability analysis \cite{8263977} \cite{https://doi.org/10.48550/arxiv.2106.13176} and artificial potential fields \cite{Singletary2021ComparativeAO}.
To obtain formal guarantees on safety (e.g., collision avoidance with obstacles), a safety critical control algorithm encompassing the trajectory tracking/planning algorithm is required that prioritizes safety over tracking. Control Barrier Functions \cite{Ames_2017} (CBFs) based approach is one such strategy in which a safe state set defined by inequality constraints is designed for the vehicle, and its quadratic programming (QP) formulation ensures forward invariance of these sets for all time.

A prime advantage of using CBF-based quadratic programs is that they work efficiently on real-time practical applications; that is, optimal control inputs can be computed at a very high frequency on off-the-shelf electronics. It can be applied as a fast safety filter over existing state-of-the-art path planning /trajectory tracking /obstacle avoiding controllers\cite{9682803}.
They are already being used for different behaviors in UGV's and self-driving cars like Adaptive Cruise Control  (ACC) \cite{7040372}, lane changing \cite{https://doi.org/10.48550/arxiv.2103.12382}, obstacle avoidance \cite{7864310} \cite{9029446} \cite{9112342} and roundabout crossings \cite{9565037}.
However, two major challenges remain that prevent the successful deployment of these CBF-QPs in autonomous systems: a) Existing CBFs are not able to handle the nonholonomic nature of autonomous vehicles well. They provide limited control capability in models like the unicycle and bicycle models i.e., the solutions from the CBF-QPs have either no steering or forward thrust capability under certain cases (this is shown in Section \ref{section: Background}), and b) Existing CBFs are not able to handle dynamic obstacles well i.e., the controllers are not able to avoid collisions with moving obstacles, which is also shown in Section \ref{section: Background}.

%


With the goal of addressing the above challenges, we propose a new class of CBFs via the collision cone approach. In particular, we generate a new class of constraints that ensure that the relative velocity between the obstacle and the vehicle always points away from the direction of the vehicle's approach. Assuming ellipsoidal shape for the obstacles \cite{709600}, the resulting set of unwanted directions for potential collision forms a conical shape, giving rise to the synthesis of \textbf{Collision Cone Control Barrier Functions (C3BFs)}. The C3BF-based QP optimally calculates inputs such that the relative velocity vector direction is kept out of the collision cone (unsafe set) for all time. The approach is incorporated and demonstrated using the acceleration-controlled unicycle and bicycle models. 
\label{section: Literature Review}



The idea of collision cones was first introduced in \cite{Fiorini1993, doi:10.1177/027836499801700706, 709600} as a means to geometrically represent the possible set of velocity vectors of the vehicle that lead to collision. 
The approach was extended for irregularly shaped robots and obstacles with unknown trajectories both in 2D \cite{709600} and 3D space \cite{Chakravarthy2012}. 
This was commonly used for aerospace applications like missile guidance \cite{doi:10.2514/1.G005879} and conflict detection and resolution in aircraft \cite{doi:10.2514/6.2004-4879}. This was further extended to Model Predictive Control by formulating the cones as constraints \cite{babu2018model}.
In our work, we combine the collision cone approach with CBFs to develop a novel obstacle-avoiding algorithm for UGVs. With the benefits of both CBFs and the Collision cone model, our C3BF algorithm is computationally efficient, works well in both static \& dynamic environments, and guarantees safety. We will demonstrate this in simulation. 



\section{Background}
\label{section: Background}
In this section, we provide the relevant background necessary to formulate our problem of moving obstacle avoidance. Specifically, we first describe the vehicle models considered in our work; namely the unicycle and the bicycle models. Next, we formally introduce Control Barrier Functions (CBFs) and their importance for real-time safety-critical control for these types of vehicle models. Finally, we explain the shortcomings in existing CBF approaches in the context of collision avoidance of moving obstacles.

\subsection{Vehicle models}
\subsubsection{Acceleration controlled unicycle model}
\par A unicycle model has state variables $x_p$, $y_p$, $\theta, v, \omega$ denoting the pose, linear velocity, and angular velocity, respectively. The control inputs are linear acceleration $(a)$ and angular acceleration $(\alpha)$. In Fig. \ref{fig:models} we show a differential drive robot, which is modeled as a unicycle. The resulting dynamics of this model is shown below:
\begin{figure}
    \centering
    \includegraphics[width=0.4\linewidth]{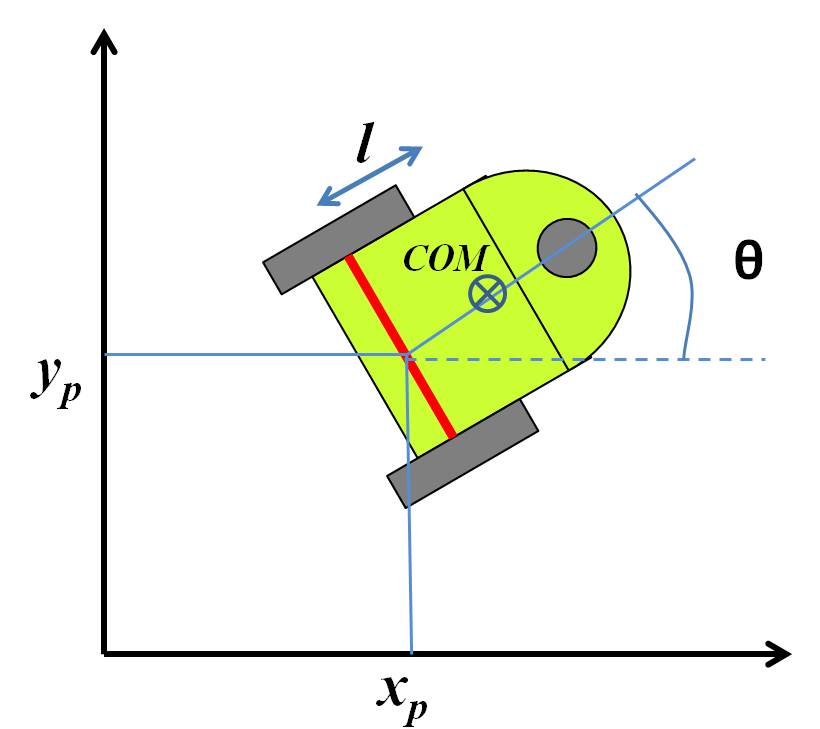}
    \includegraphics[width=0.52\linewidth]{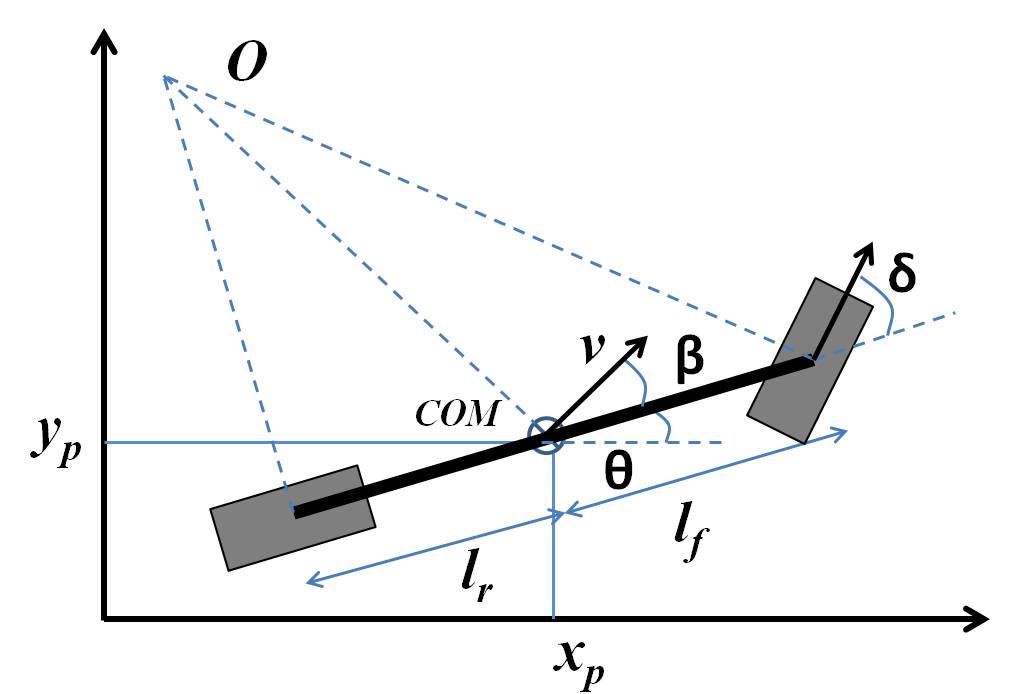}
\caption{Schematic of Unicycle (left); Bicycle model (right).}
\label{fig:models}
\end{figure}

\begin{equation}
	\begin{bmatrix}
		\dot{x}_p \\
		\dot{y}_p \\
		\dot{\theta} \\
		\dot{v} \\
		\dot{\omega}
	\end{bmatrix}
	=
        \begin{bmatrix}
            v\cos\theta\\
            v\sin\theta\\
            \omega \\
            0 \\
            0
        \end{bmatrix}
	+
	\begin{bmatrix}
            0 & 0 \\
            0 & 0 \\
            0 & 0 \\
            1 & 0 \\
            0 & 1
	\end{bmatrix}
	\begin{bmatrix}
		a \\
		\alpha
	\end{bmatrix}
    \label{eqn:Acceleration controlled Unicycle model}
\end{equation}

While the commonly used unicycle model in literature includes linear and angular velocities $v,\omega$ as inputs, we use accelerations as inputs. This is due to the fact that differential drive robots have torques as inputs to the wheels that directly affect acceleration. In other words, we can treat the force/acceleration applied from the wheels as inputs. As a result, $v,\omega$ become state variables in our model.

\subsubsection{Acceleration controlled bicycle model}
The bicycle model has two wheels, where the front wheel is used for steering (see Fig. \ref{fig:models}). This model is typically used for self-driving cars,
where we treat the front and rear wheel sets as a single virtual wheel (for each set) by considering the difference of steer in right and left wheels to be negligible 
\cite{https://doi.org/10.48550/arxiv.2103.12382, Rajamani2014-jb, 7995816}. The bicycle dynamics are as follows:
\begin{align}
    \begin{bmatrix}
        \dot x_p \\
        \dot y_p \\
        \dot \theta \\
        \dot v 
    \end{bmatrix} 
    & = 
    \begin{bmatrix}
        v \cos (\theta + \beta) \\
        v \sin (\theta + \beta) \\
        \frac{v}{l_r} \sin (\beta) \\
        a 
    \end{bmatrix},
    \label{eq:bicyclemodel} \\
	\text{where} \quad \beta &= \tan^{-1}\left(\frac{l_r}{l_f + l_r}\tan(\delta)\right), \label{eq:SlipSteeringConv}
\end{align}
$x_p$ and $y_p$ denote the coordinates of the vehicle’s center of mass (CoM) in an inertial frame. $\theta$ represents the orientation of the vehicle with respect to the $x$ axis. $a$ is the linear acceleration at CoM.
$l_f$ and $l_r$ are the distances of the front and rear axles from the CoM,  respectively.
$\delta$ is the steering angle of the vehicle and 
$\beta$ is the vehicle's slip angle, i.e., the steering angle of the vehicle mapped to its CoM (see Fig. \ref{fig:models}).
This is not to be confused with the tire slip angle.

\begin{remark}
Similar to \cite{https://doi.org/10.48550/arxiv.2103.12382}, we assume that the slip angle is constrained to be small. As a result, we approximate $\cos \beta \approx 1$ and $\sin \beta \approx \beta$. Accordingly, we get the following simplified dynamics of the bicycle model:
\begin{equation}
\label{eqn:bicyclemodel with small beta}
	\underbrace{\begin{bmatrix}
		\dot{x}_p \\
		\dot{y}_p \\
		\dot{\theta} \\
		\dot{v}
	\end{bmatrix}}_{\dot{\state}}
	=
	\underbrace{\begin{bmatrix}
		v \cos\theta \\
		v \sin \theta \\
		0 \\
		0
	\end{bmatrix}}_{f(\state)}
	+
	\underbrace{\begin{bmatrix}
		0 & - v\sin\theta \\
		0 & v\cos\theta \\
		0 & \frac{v}{l_r} \\
		1 & 0
	\end{bmatrix}}_{g(\state)}
	\underbrace{\begin{bmatrix}
		a \\
		\beta
	\end{bmatrix}}_{u}.
\end{equation}
Since the control inputs $a, \beta$ are now affine in the dynamics, CBF-based Quadratic Programs (CBF-QPs) can be constructed directly to yield real-time control laws, as explained next.
\end{remark}

\subsection{Control barrier functions (CBFs)}
Having described the vehicle models, we now formally introduce Control Barrier Functions (CBFs) and their applications in the context of safety. 
%
Consider a nonlinear control system in affine form:
\begin{equation}
	\dot{\state} = f(\state) + g(\state)u
	\label{eqn: affine control system}
\end{equation}
where $\state \in \mathcal{D} \subseteq \mathbb{R}^n$ is the state of system, and $u \in \mathbb{U} \subseteq \mathbb{R}^m$ the input for the system. Assume that the functions $f: \mathbb{R}^n \rightarrow \mathbb{R}^n$ and $g: \mathbb{R}^n \rightarrow \mathbb{R}^{n \times m}$ are continuously differentiable. Given a Lipschitz continuous control law $u = k(\state)$, the resulting closed loop system $\dot{\state} = f_{cl}(\state) = f(\state) + g(\state)k(\state)$ yields a solution $\state(t)$, with initial condition $\state(0) = \state_0$.
%
Consider a set $\mathcal{C}$ defined as the \textit{super-level set} of a continuously differentiable function $h:\mathcal{D}\subseteq \mathbb{R}^n \rightarrow \mathbb{R}$ yielding,
\begin{align}
\label{eq:setc1}
	\mathcal{C}                        & = \{ \state \in \mathcal{D} \subset \mathbb{R}^n : h(\state) \geq 0\} \\
\label{eq:setc2}
	\partial\mathcal{C}                & = \{ \state \in \mathcal{D} \subset \mathbb{R}^n : h(\state) = 0\}\\
\label{eq:setc3}
	\text{Int}\left(\mathcal{C}\right) & = \{ \state \in \mathcal{D} \subset \mathbb{R}^n : h(\state) > 0\}
\end{align}
It is assumed that $\text{Int}\left(\mathcal{C}\right)$ is non-empty and $\mathcal{C}$ has no isolated points, i.e. $\text{Int}\left(\mathcal{C}\right) \neq \phi$ and $\overline{\text{Int}\left(\mathcal{C}\right)} = \mathcal{C}$. 
The system is safe w.r.t. the control law $u = k(\state)$ if
	$\forall \: \state(0) \in \mathcal{C} \implies \state(t) \in \mathcal{C} \;\;\; \forall t \geq 0$.
We can mathematically verify if the controller $k(\state)$ is safeguarding or not by using Control Barrier Functions (CBFs), which is defined next.

\begin{definition}[Control barrier function (CBF)]{\it
\label{definition: CBF definition}
Given the set $\mathcal{C}$ defined by \eqref{eq:setc1}-\eqref{eq:setc3}, with $\frac{\partial h}{\partial \state}(\state) \neq 0\; \forall \state \in \partial \mathcal{C}$, the function $h$ is called the control barrier function (CBF) defined on the set $\mathcal{D}$, if there exists an extended \textit{class} $\mathcal{K}$ function $\kappa$ such that for all $\state \in \mathcal{D}$:

\begin{equation}
\begin{aligned}
    \underbrace{\text{sup}}_{ u \in \mathbb{U}}\! \left[\underbrace{\mathcal{L}_{f} h(\state) + \mathcal{L}_g h(\state)u} \iffalse+ \frac{\partial h}{\partial t}\fi_{\dot{h}\left(\state, u\right)} \! + \kappa\left(h(\state)\right)\right] \! \geq \! 0
\end{aligned}
\end{equation}
where $\mathcal{L}_{f} h(\state) = \frac{\partial h}{\partial \state}f(\state)$ and $\mathcal{L}_{g} h(\state)= \frac{\partial h}{\partial \state}g(\state)$ are the Lie derivatives. 
}
\end{definition}

Given this definition of a CBF, we know from \cite{Ames_2017} and \cite{8796030} that any Lipschitz continuous control law $k(\state)$ satisfying the inequality: $\dot{h} + \kappa( h )\geq 0$ ensures safety of $\mathcal{C}$ if $x(0)\in \mathcal{C}$, and asymptotic convergence to $\mathcal{C}$ if $x(0)$ is outside of $\mathcal{C}$. 
It is worth mentioning that CBFs can be also defined just on $\mathcal{C}$, wherein we ensure only safety. This will be useful for the bicycle model \eqref{eqn:bicyclemodel with small beta}, which is described later on.

\subsection{Controller synthesis for real-time safety}
Having described the CBF and its associated formal results, we now discuss its Quadratic Programming (QP) formulation. 
CBFs are typically regarded as \textit{safety filters} which take the desired input (reference controller input) $u_{ref}(\state,t)$ and modify this input in a minimal way: 

\begin{equation}
\begin{aligned}
\label{eqn: CBF QP}
u^{*}(x,t) &= \min_{u \in \mathbb{U} \subseteq \mathbb{R}^m} \norm{u - u_{ref}(x,t)}^2\\
\quad & \textrm{s.t. } \mathcal{L}_f h(x) + \mathcal{L}_g h(x)u + \kappa \left(h(x)\right) \geq 0\\
\end{aligned}
\end{equation}
This is called the Control Barrier Function based Quadratic Program (CBF-QP). If $\mathbb{U}=\mathbb{R}^m$, then the QP is feasible, and the explicit solution is given by
\begin{equation*}
	u^{*}(x, t) = u_{ref}(x, t) + u_{safe}(x,t)
\end{equation*}
where $u_{safe}(x,t)$ 
is given by
\begin{multline}\label{eq:CBF-QP}
u_{safe}(x, t) \!=\!
	\begin{cases}
		0 & \text{for } \psi(x, t) \geq 0 \\
		-\frac{\mathcal{L}_{g}h(x)^T \psi(x, t)}{\mathcal{L}_{g}h(x)\mathcal{L}_{g}h(x)^T} & \text{for } \psi(x, t) < 0
	\end{cases}
\end{multline}
where $\psi (x,t) := \dot{h}\left(x, u_{ref}(x, t)\right) + \kappa \left(h(x)\right)$. The sign change of $\psi$ yields a switching type of a control law.

\subsection{Classical CBFs and moving obstacle avoidance}
Having introduced CBFs, we now explore collision avoidance in unmanned ground vehicles (UGVs). In particular, we discuss the problems associated with the classical CBF-QPs, especially with the velocity obstacles. 
We also summarize and compare with C3BFs in Table \ref{table: types of CBFs}.

\begin{table*}[t]
\begin{center}
    \begin{threeparttable}
	\begin{tabular}{| c | c | c | c |}
		\hline
		\textbf{CBFs} & \textbf{Vehicle Models} & \textbf{Static Obstacle} $(c_x, c_y)$ & \textbf{Moving Obstacle} $(c_x(t), c_y(t)) ^\dag$ \\
		\hline
		\textbf{Ellipse CBF}  & Acceleration controlled unicycle \eqref{eqn:Acceleration controlled Unicycle model}   & Not a valid CBF                        & Not a valid CBF                               \\
		\hline
		\textbf{Ellipse CBF}  & Bicycle \eqref{eqn:bicyclemodel with small beta}    & Valid CBF, No acceleration                        & Not a valid CBF                               \\
		\hline
		\textbf{C3BF}   & Acceleration controlled unicycle \eqref{eqn:Acceleration controlled Unicycle model} & Valid CBF in $\mathcal{D}$                       & Valid CBF in $\mathcal{D}$                              \\
		\hline
            \textbf{C3BF} & Bicycle \eqref{eqn:bicyclemodel with small beta} & Valid CBF in $\mathcal{C}$ & Valid CBF in $\mathcal{C}$ \\
            \hline
	\end{tabular}
	\begin{tablenotes}\footnotesize
    \item[$\dag$] $(c_x(t), c_y(t))$ are continuous (or at least piece-wise continuous) functions of time
    \end{tablenotes}
	\end{threeparttable}
\end{center}
\caption{Comparison between the CBFs: Ellipse CBF \eqref{eqn:Ellipse-CBF}, and the proposed C3BF for different vehicle models.}
\label{table: types of CBFs}
\end{table*}

\subsubsection{Ellipse-CBF Candidate - Unicycle}
Consider the following CBF candidate:
\begin{equation}
    h(\state,t) = \left(\frac{c_x(t) - x_p}{c_1}\right)^2 + \left(\frac{c_y(t) - y_p}{c_2}\right)^2 - 1,
    \label{eqn:Ellipse-CBF}
\end{equation}
which approximates an obstacle with an ellipse with center $(c_x(t), c_y(t))$ and axis lengths $c_1,c_2$. We assume that $c_x(t),c_y(t)$ are differentiable and their derivatives are piece-wise constants. 
Since $h$ in \eqref{eqn:Ellipse-CBF} is dependent on time (e.g. moving obstacles), the resulting set $\mathcal{C}$ is also dependent on time. To analyze this class of sets, time-dependent versions of CBFs can be used \cite{IGARASHI2019735}. Alternatively, we can reformulate our problem to treat the obstacle position $c_x,c_y$ as states, with their derivatives being constants. This will allow us to continue using the classical CBF given by Definition \ref{definition: CBF definition} including its properties on safety. The derivative of \eqref{eqn:Ellipse-CBF} is
\begin{align}
\frac{2(c_x-x)(\dot c_x - v\cos \theta)}{c_1^2} +\frac{2(c_y-y)(\dot c_y - v \sin \theta)}{c_2^2},
\end{align}
which has no dependency on the inputs $a, \alpha$. Hence, $h$ will not be a valid CBF for the acceleration-based model \eqref{eqn:Acceleration controlled Unicycle model}. 
However, for static obstacles,
if we choose to use the velocity model (with $v,\omega$ as inputs instead of $a,\alpha$), 
then $h$ will certainly be a valid CBF, but the vehicle will have limited control capability i.e., it loses steering $\omega$.

\subsubsection{Ellipse-CBF Candidate - Bicycle}
For the bicycle model, the derivative of $h$ \eqref{eqn:Ellipse-CBF} yields
\begin{align}
    \dot h = & 2 (c_x - x_p) ( \dot c_x - v \cos \theta + v (\sin \theta) \beta)/c_1^2 \nonumber \\
    & + 2 (c_y - y_p) ( \dot c_y - v \sin \theta - v (\cos \theta) \beta)/c_2^2,
\end{align}
which only has $\beta$ as the input. Furthermore, the derivatives $\dot c_x, \dot c_y$ are free variables i.e., the obstacle velocities can be selected in such a way that the constraint $\dot h(x,u)+ \kappa (h(x)) < 0$, whenever $\mathcal{L}_g h=0$. This implies that $h$ is not a valid CBF for moving obstacles.

It is worth mentioning that for the acceleration-controlled unicycle models \eqref{eqn:Acceleration controlled Unicycle model}, we can use 
another class of CBFs introduced specifically for constraints with higher relative degrees \cite{7524935,https://doi.org/10.48550/arxiv.1903.04706}, \cite{9456981}. However, our goal in this paper is to develop a generic CBF formulation that provides safety guarantees in both the unicycle and bicycle models. We propose this next.

\section{Collision Cone CBF (C3BF)}
\label{section: Collision Cone CBF}
Having described the shortcomings of existing approaches for collision avoidance, we will now describe the proposed method i.e., the collision cone CBFs (C3BFs).
A collision cone, defined for a pair of objects, is a set that can be used to predict the possibility of collision between the two objects based on the direction of their relative velocity. The collision cone of an object pair represents the directions, which if traversed by either object, will result in a collision between the two. 
We will treat the obstacles as ellipses with the vehicle reduced to a point; therefore, throughout the rest of the paper, the term collision cone will refer to this case with the ego-vehicle's center being the point of reference. 






Consider an ego-vehicle defined by the system (\ref{eqn: affine control system}) and a moving obstacle (pedestrian, another vehicle, etc.). This is shown pictorially in Fig. \ref{Fig:Construction of Collision Cone}. We define the velocity and positions of the obstacle w.r.t. the ego vehicle. We \textit{over-approximated} the obstacle to be an ellipse and draw two tangents from the vehicle's center to a conservative circle encompassing the ellipse, taking into account the ego-vehicle's dimensions ($r = max(c_1, c_2) + \frac{Width_{vehicle}}{2}$). 
For a collision to happen, the relative velocity of the obstacle must be pointing towards the vehicle. Hence, the relative velocity vector must not be pointing into the pink shaded region EHI in Fig. \ref{Fig:Construction of Collision Cone}, which is a cone. 
%
%
Let $\mathcal{C}$ be this set of safe directions for this relative velocity vector. If there exists a function $h:\mathcal{D}\subseteq \mathbb{R}^n \rightarrow \mathbb{R}$ satisfying \textit{Definition: \ref{definition: CBF definition}} on $\mathcal{C}$, then we know that a Lipshitz continuous control law obtained from the resulting QP (\ref{eqn: CBF QP}) for the system ensures that the vehicle won't collide with the obstacle even if the reference $u_{ref}$ tries to direct them towards a collision course. 
%
%
%
This novel approach of avoiding the pink cone region
gives rise to \textbf{Collision Cone Control Barrier Functions (C3BFs)}. 

\begin{figure}
    \centering
    \begin{tikzpicture}[
      collisioncone/.style={shape=rectangle, fill=red, line width=2, opacity=0.30},
      obstacleellipse/.style={shape=rectangle, fill=blue, line width=2, opacity=0.35},
    ]
        
        \def\r{1.32003}; 
        \def\q{-3.5}; 
        \def\x{{\r^2/\q}}; 
        \def\y{{\r*sqrt(1-(\r/\q)^2}}; 
        \def\z{{\q - abs(\q - (\r^2/\q))}};
        \coordinate (Q) at (\q,0); 
        \coordinate (P) at (\x,\y); 
        \coordinate (O) at (0.0, 0); 
        \coordinate (E) at (\q, 0); 
        \coordinate (K) at (\x, {-\y}); 
        \coordinate (H) at (\z, \y);
        \coordinate (I) at (\z, {-\y});
        
        \draw[name path = aux, red!60, very thick, dashed] (O) circle (1.32003);
        \draw[blue!50, thick, fill=blue!20] (O) ellipse (1.20 and 0.55);
        \draw[black, thick] (E) -- (O) node [midway, below] {$\|\prel\|$};
        
        
        \draw[black, thick, name path = tangent] ($(Q)!-0.0!(P)$) -- ($(Q)!1.3!(P)$); 
        \draw[black, thick, name path = normal] ($(O)!-0.0!(P)$) -- ($(O)!1.4!(P)$);
        \draw[black, thick] ($(Q)!-0.0!(K)$) -- ($(Q)!1.3!(K)$);
        
        \draw[black, thick, name path = tangent, dashed] ($(Q)!-0.0!(H)$) -- ($(Q)!1.1!(H)$);
        \draw[black, thick, dashed] ($(Q)!-0.0!(I)$) -- ($(Q)!1.1!(I)$);
        
        
        \tkzMarkRightAngle[draw=black,size=.2](O,P,Q);
        \tkzMarkAngle[draw=black, size=0.75](O,Q,P);
        \tkzLabelAngle[dist=1.0](O,Q,P){$\phi$};
        
        \path[shade, left color=red, right color = red, opacity=0.2] (E) -- (H) -- (I) -- cycle;
        
        \fill [black] (E) circle (2pt) node[anchor=north, black] (n1) {$(x,y)$} node[anchor=south, black] (n1) {E};
        \fill [blue] (O) circle (2pt) node[anchor=north, blue] (n1) {$(c_x, c_y)$} node[anchor=south east, blue] (n1) {O}; 
        \fill [black] (P) circle (2pt) node[anchor=south, black] (n1) {$P$};
        \fill [black] (K) circle (2pt) node[anchor=north, black] (n1) {$K$};
        \fill [black] (H) circle (2pt) node[anchor=south, black] (n2) {$H$};
        \fill [black] (I) circle (2pt) node[anchor=north, black] (n1) {$I$};
        
        \draw [<->, color=black, thick, dashed] ([xshift=5 pt, yshift=0 pt]O) -- ([xshift=5 pt, yshift=0 pt]P) node [midway, right] {$r = a+\frac{w}{2}$};
        \draw [<->, color=black, thick, dashed] (O) -- (1.20, 0) node [midway, above] {$a$};
        
        \matrix [above right,nodes in empty cells, matrix of nodes, column sep=0.5cm, inner sep=6pt] at (current bounding box.north west) {
          \node [collisioncone,label=right:{\footnotesize Collision Cone}] {}; &
          \node [obstacleellipse,label=right:{\footnotesize Obstacle Ellipse}] {}; \\
        };
    \end{tikzpicture}
    \caption{Construction of collision cone for an elliptical obstacle considering the ego vehicle's dimensions (width: $w$).} 
    \label{Fig:Construction of Collision Cone}
\end{figure}
\subsection{Application to systems}
\subsubsection{Acceleration controlled unicycle model}
\label{section: acc controlled unicycle model cccbf}
We first obtain the relative position vector between the body center of the unicycle and the center of the obstacle. 
Therefore, we have
\begin{align}\label{eq:positionvectorunicycle}
    \prel := \begin{bmatrix}
        c_x - (x_p + l \cos(\theta)) \\
        c_y - (y_p + l \sin(\theta))
    \end{bmatrix}
\end{align}
Here $l$ is the distance of the body center from the differential drive axis (see Fig. \ref{fig:models}). We obtain its velocity as
\begin{align}\label{eq:velocityvectorunicycle}
    \vrel := \begin{bmatrix}
        \dot c_x - (v \cos (\theta) - l \sin(\theta)*\omega) \\
        \dot c_y - (v \sin (\theta) + l \cos(\theta)*\omega)
    \end{bmatrix}.
\end{align}
We propose the following CBF candidate:
%
\begin{equation}
    h(\state, t) = <\prel,\vrel> + \|\prel\|\|\vrel\|\cos\phi
    \label{eqn:CC-CBF}
\end{equation}
where, $\phi$ is the half angle of the cone, the expression of $\cos\phi$ is given by $\frac{\sqrt{\|\prel\|^2 - r^2}}{\|\prel\|}$ (see Fig. \ref{Fig:Construction of Collision Cone}). 
The constraint simply ensures that the angle between $\prel, \vrel$ is less than $180^\circ - \phi$.  
We have the following first result of the paper:
%
\begin{theorem}\label{thm:unicycletheorem}{\it
Given the acceleration controlled unicycle model \eqref{eqn:Acceleration controlled Unicycle model}, the proposed CBF candidate \eqref{eqn:CC-CBF} with $\prel,\vrel$ defined by \eqref{eq:positionvectorunicycle}, \eqref{eq:velocityvectorunicycle} is a valid CBF defined for the set $\mathcal{D}$.}
\end{theorem}
\begin{proof}
Taking the derivative of \eqref{eqn:CC-CBF} yields ($\vrel \neq 0$)
\begin{align}
\dot h = &  < \preldot, \vrel > + < \prel, \vreldot >  \nonumber \\
 & + < \vrel, \vreldot > \frac{\sqrt{\|\prel\|^2 - r^2}}{\|\vrel\|} \nonumber \\
 & + < \prel, \preldot > \frac{\|\vrel\| }{\sqrt{\|\prel\|^2 - r^2}}.
 \label{eqn:h_derivative}
\end{align}
Further $\preldot  = \vrel$ and
\begin{align*}
    \vreldot = \begin{bmatrix}
        - a \cos \theta + v (\sin \theta) \omega + l (\cos \theta) \omega^2 + l (\sin \theta) \alpha \\
        -a \sin \theta - v (\cos \theta) \omega + l (\sin \theta) \omega^2 - l (\cos \theta) \alpha
    \end{bmatrix}. \nonumber
\end{align*}
Given $\vreldot$ and $\dot h$, we have the following expression for $\mathcal{L}_g h$:
\begin{align}
    \mathcal{L}_g h = \begin{bmatrix}
        < \prel + \vrel \frac{\sqrt{\|\prel\|^2 - r^2}}{\|\vrel\|}, \begin{bmatrix}
            - \cos \theta \\
            - \sin \theta
        \end{bmatrix}>  \\
                < \prel + \vrel \frac{\sqrt{\|\vrel\|^2 - r^2}}{\|\vrel\|}, \begin{bmatrix}
            l \sin \theta \\
            - l \cos \theta
        \end{bmatrix}> 
    \end{bmatrix}^T,
\end{align}
It can be verified that for $\mathcal{L}_gh$ to be zero, we can have the following scenarios:
\begin{itemize}
    \item $\prel + \vrel \frac{\sqrt{\|\prel\|^2 - r^2}}{\|\vrel\|}=0$, which is not possible. Firstly, $\prel=0$ indicates that the vehicle is already inside the obstacle. Secondly, if the above equation were to be true for a non-zero $\prel$, then $\vrel/\|\vrel\| = - \prel/\sqrt{\|\prel\|^2 - r^2}$. This is also not possible as the magnitude of LHS is $1$, while that of RHS is $>1$.
    \item $\prel + \vrel \frac{\sqrt{\|\vrel\|^2 - r^2}}{\|\vrel\|}$ is perpendicular to both $\begin{bmatrix}
            - \cos \theta \\
            - \sin \theta
        \end{bmatrix}$ and  $\begin{bmatrix}
            l \sin \theta \\
            - l \cos \theta
        \end{bmatrix}$, which is also not possible.
\end{itemize}
This implies that $\mathcal{L}_gh$ is always a non-zero matrix, implying that $h$ is a valid CBF.
\end{proof}
\begin{remark}
{\it
Since $\mathcal{L}_g h \neq 0$, we can infer from \cite[Theorem 8]{XU201554} that the resulting QP given by \eqref{eq:CBF-QP} is Lipschitz continuous. Hence, we can construct CBF-QPs with the proposed CBF \eqref{eqn:CC-CBF} for the unicycle model and guarantee collision avoidance. In addition,  if $h(x(0))<0$, then we can construct a class $\mathcal{K}$ function $\kappa$ in such a way that the magnitude of $h$ exponentially decreases over time, thereby minimizing the violation. We will demonstrate these scenarios in 
Section \ref{section: Results and Discussions}.
}
\end{remark}
\subsubsection{Acceleration controlled Bicycle model}
\label{section: bicycle model cccbf}
For the approximated bicycle model \eqref{eqn:bicyclemodel with small beta}, we define the following:
\begin{align}\label{eq:positionvectorbicycle}
    \prel := &
    \begin{bmatrix}
        c_x - x_p &       c_y - y_p 
    \end{bmatrix}^T \\
\label{eq:velocityvectorbicycle}
 \vrel := & \begin{bmatrix}
        \dot c_x - v \cos \theta &        \dot c_y - v \sin \theta 
    \end{bmatrix}^T,
\end{align}
Here $\vrel$ is NOT equal to the relative velocity $\preldot$. However, for small $\beta$, we can assume that $\vrel$ is the difference between obstacle velocity and the velocity component along the length of the vehicle $v \cos \beta \approx v$. In other words, the goal is to ensure that this approximated velocity $\vrel$ is pointing away from the cone. This is an acceptable approximation as $\beta$ is small and the obstacle radius chosen was conservative (see Fig. \ref{Fig:Construction of Collision Cone}). 
We have the following result:
\begin{theorem}\label{thm:bicycletheorem}{\it
Given the bicycle model \eqref{eqn:bicyclemodel with small beta}, the proposed candidate CBF \eqref{eqn:CC-CBF} with $\prel,\vrel$ defined by \eqref{eq:positionvectorbicycle}, \eqref{eq:velocityvectorbicycle} is a valid CBF defined for the set $\mathcal{C}$.}
\end{theorem}
\begin{proof}
We need to show that $\mathcal{L}_g h = 0$ $\implies$ $\dot h + \kappa(h)\geq0$. The derivative of $h$ \eqref{eqn:CC-CBF} yields
\eqref{eqn:h_derivative}. Further using \eqref{eqn:bicyclemodel with small beta}: $\preldot = \vrel + \beta[ v \sin \theta, - v \cos \theta]^T$ and 
\begin{align}
    \vreldot = \begin{bmatrix} 
        -  \cos \theta &  v\sin \theta  \\
        -  \sin \theta & - v \cos \theta 
    \end{bmatrix}\begin{bmatrix}
        a \\ \frac{v}{l_r}\beta
    \end{bmatrix}.
\end{align}
When $\mathcal{L}_g h=0$, we have 
\begin{equation}
    \dot h + \kappa (h) = < \vrel, \vrel > +  \frac{< \prel, \vrel >\|\vrel\| }{\sqrt{\|\prel\|^2 - r^2}} + \kappa (h). \nonumber
\end{equation}
Rewriting the above equation (when $\vrel \neq 0$) yields
\begin{equation}
    \frac{\|\vrel\|}{\sqrt{\|\prel\|^2 - r^2}}\left( h + \frac{ \sqrt{\|\prel\|^2 - r^2}}{\|\vrel\|}\kappa (h)
    \right).
\end{equation}  
Since $\sqrt{\|\prel\|^2 - r^2}$ and $\|\vrel\|$ are positive quantities, the entire quantity above is $\geq 0$ for all $x\in \mathcal{C}$. We can note that $\vrel = 0$ means either both the vehicle and obstacle are stationary or both are moving parallel to each other in the same direction. In both cases, C3BF will never activate as the vehicle will not be on the course of the collision. This completes the proof. 
\end{proof}
\begin{remark}
{\it
Theorem \ref{thm:bicycletheorem} is different from Theorem \ref{thm:unicycletheorem} as the CBF inequality is satisfied in the set $\mathcal{C}$ and not in $\mathcal{D}$. In other words, forward invariance of $\mathcal{C}$ can be guaranteed, but not asymptotic convergence of $\mathcal{C}$. However, modifications of the control formulation are possible to extend the result for $\mathcal{D}$, which will be a subject of future work.
}
\end{remark}

\section{Results and Discussions}
\label{section: Results and Discussions}
\par In this section, we provide the simulation results to validate the proposed C3BF-QP. All the simulations are done in Pybullet using TurtleBot3 (modeled as a unicycle), and F1-tenth (modeled as a bicycle). 


\begin{figure}
        \begin{minipage}{0.21\textwidth} 
        \begin{subfigure}{\linewidth}
        \includegraphics[width=\linewidth]{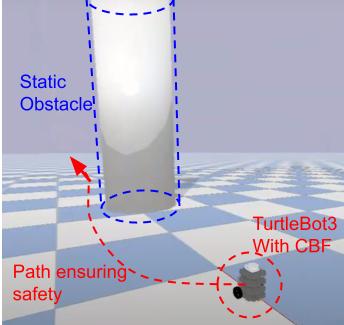}
        \caption{}
        \end{subfigure}
        \addtocounter{subfigure}{1} 
        \begin{subfigure}{\linewidth}
        \includegraphics[width=\linewidth]{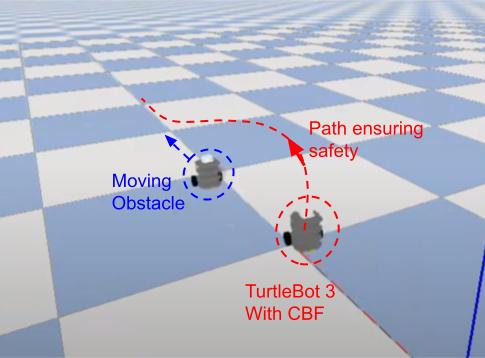}
        \caption{}
        \end{subfigure}
        \end{minipage} 
        \begin{minipage}{0.27\textwidth} 
        \addtocounter{subfigure}{-2} 
        \begin{subfigure}{\linewidth}
        \includegraphics[width=\linewidth]{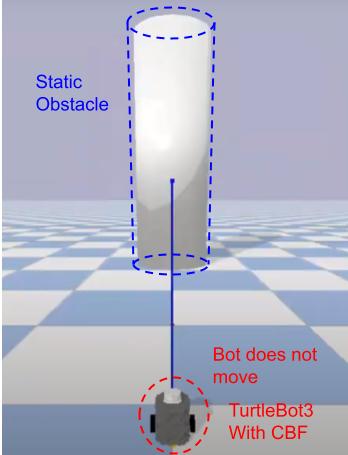}
        \caption{}
        \end{subfigure}
        \end{minipage} 
        \addtocounter{subfigure}{1}
        \begin{subfigure}[b]{0.48\textwidth}
        \includegraphics[width=\textwidth]{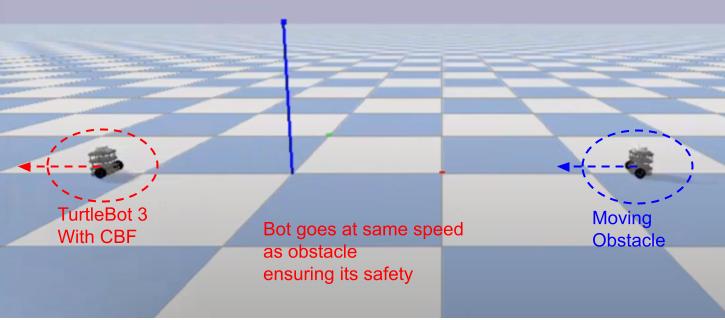}
        \caption{}
        \end{subfigure}

\caption{Interaction of TurtleBot3 (Unicycle Model) with static obstacles: (a), (b), and moving obstacles (c), (d)}
\label{fig:Unicycle-obs}
\end{figure}

\begin{figure}
       \centering
        \begin{subfigure}[b]{0.23\textwidth}
        \includegraphics[width=\textwidth]{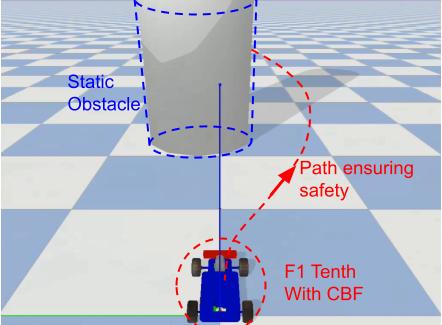}
        \caption{}
        \end{subfigure}
        \begin{subfigure}[b]{0.23\textwidth}
        \includegraphics[width=\textwidth]{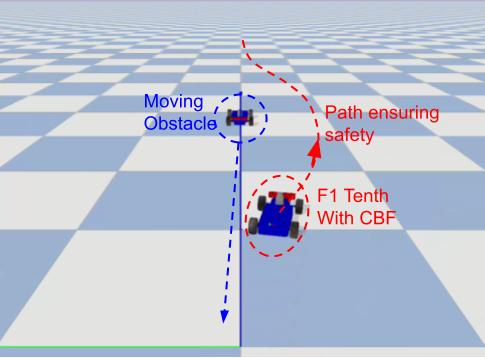}
        \caption{}
        \end{subfigure}
        \begin{subfigure}[b]{0.46\textwidth}
        \includegraphics[width=\textwidth]{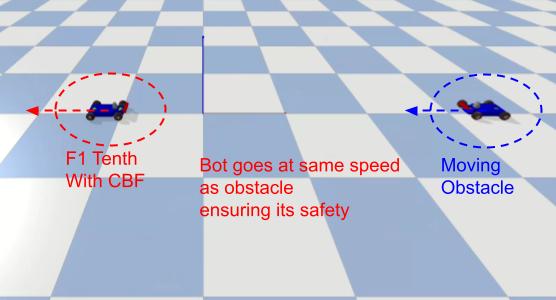}
        \caption{}
        \end{subfigure}
        \caption{Interaction of F1 Tenth (Bicycle) with static obstacles: (a),  and moving obstacles (b),(c).}
        \label{fig:Bicycle-obs}
\end{figure}

\subsection{Acceleration Controlled Unicycle Model}

We have considered the reference control inputs as a simple PD controller for simulating which can be replaced with any state-of-the-art path planning/ trajectory tracking/obstacle-avoiding controller like Stanley controller \cite{4282788}.
Constant target velocities were chosen to verify the C3BF-QP. For the class $\mathcal{K}$ function in the CBF inequality, we chose $\kappa(h) = \gamma h$, where $\gamma=1$. Fig. \ref{fig:Unicycle-obs} (a) shows the
moving around (the obstacle) and (b) shows braking (in case the bot and obstacle share the same axis)  while interacting with the static obstacle. Fig. \ref{fig:Unicycle-obs}(c) shows overtaking and (d) shows the reversing behavior (in case the obstacle and the bot shares the axis) while interacting with the moving obstacle.

\subsection{Acceleration Controlled Bicycle Model}
We have extended and validated our C3BF algorithm for the bicycle model \eqref{eqn:bicyclemodel with small beta} which is a good approximation of actual car dynamics under the assumption of small lateral acceleration ($\leq 0.5\mu g$, $\mu$ is the friction co-efficient) \cite{https://doi.org/10.48550/arxiv.2103.12382} \cite{7995816}.  
The linear acceleration reference control $a_{ref}$ was obtained from a PD controller tracking the desired velocity, and the steering reference $\beta_{ref}$ was obtained from the Stanley controller \cite{4282788}. The reference controllers were integrated with the C3BF-QP, and applied to the robot simulated on F1 Tenth in Pybullet (Fig. \ref{fig:Bicycle-obs}) with behaviors similar to the one in the Unicycle case. The resulting simulations for both Unicycle and Bicycle can be viewed in this  video\footnote{\label{note: Pybullet Simulation Video} \url{https://youtu.be/Dme7Wm9y6es}}.

\section{Conclusions}
\label{section: Conclusions}
We presented a novel CBF formulation for collision avoidance with moving obstacles, by using the concept of collision cones. 
Existing works in literature with CBFs and acceleration-controlled non-holonomic UGV models were not able to circumnavigate / brake in the presence of obstacles with non-zero velocity values. The proposed QP formulation (C3BF-QP) allows the vehicle to safely maneuver under different scenarios presented in the paper. Moreover, it can be combined as a filter with any existing state-of-the-art trajectory tracking/ path-planning/obstacle-avoiding controllers to get theoretical guarantees on safety. Future work includes experimental tests on UGVs and UAVs.

\label{section: References}
\bibliographystyle{IEEEtran}
\bibliography{references.bib}

\end{document}